# Eliminating deceptions and mistaken belief to infer conversational implicature

**Mark Lee and Yorick Wilks**
Department of Computer Science
University of Sheffield
Regent Court, 211 Portobello Street
Sheffield S1 4DP, UK
*M.Lee@dcs.shef.ac.uk*
*Y.Wilks@dcs.shef.ac.uk*

## Abstract

Conversational implicatures are usually described as being licensed by the disobeying or flouting of some principle by the speaker in cooperative dialogue. However, such work has failed to distinguish cases of the speaker flouting such a principle from cases where the speaker is either deceptive or holds a mistaken belief. In this paper, we demonstrate how the three different cases can be distinguished in terms of the beliefs ascribed to the speaker of the utterance. We argue that in the act of distinguishing the speaker's intention and ascribing such beliefs, the intended inference can be made by the hearer. This theory is implemented in ViewGen, a pre-existing belief modelling system used in a medical counselling domain.

## 1 Introduction

Previous work in pragmatics has described how speakers can communicate more than what is explicitly said by the flouting of normative principles such as those of cooperation [Grice, 1975], or the felicity conditions of an utterance's speech act [Searle, 1969]. Such theories have proved to be computationally underspecified or intractable. Underlying such principles has been a notion of truthfulness yet dialogue exchanges such as shown in (1):

(1). **System: I just deleted the subdirectory**
**Simon: David will be happy**

where the response is probably false, have received little attention in computational work. This is due to the lack of a clear distinction between cases of mistaken belief, intentional deception and ostensively deceptive but pragmatically felicitious communication. For example, in (1) Simon's reply can be regarded as ironic, deceptive or just a case of Simon holding a mistaken belief depending on whether the dialogue was cooperative, non-cooperative or possibly conflictive.

In this paper, we discuss an initial distinction between such cases with respect to the speaker's mental attitudes and show how it can be used to recognise and understand a class of conversational implicature [Grice, 1975]. This theory is implemented in ViewGen, a computer program which models the propositional attitudes of agents engaged in dialogue. ViewGen is implemented in Quintus prolog and currently we are investigating its use in a medical counselling domain where conflictive and deceptive dialogues are common.

Our theory is summarised as follows: Previous accounts of dialogue understanding have assumed that dialogues are cooperative in a Gricean sense so that the participants are truthful, informative but not verbose, relevant and clear. However, such assumptions ignore the possibility of conflicting beliefs and goals on the part of the participants.

Acts of deception and cases of mistaken belief have distinct belief conditions which the hearer can recognise. In using plan recognition to understand the meaning of a speaker's utterance, it is essential to first ascribe the correct set of beliefs to the speaker. The understanding of utterances based on either deception or mistaken beliefs is a form of keyhole recognition [Carberry, 1990] which is difficult in practice. However, both sets of belief conditions can be used in the recognition and understanding of conversational implicature in cooperative dialogue.

If the speaker is attempting to implicate some additional meaning then he or she must assume that the hearer will recognise their attempt as such. This is only possible if the speaker is sure that the hearer can eliminate the possibility of deception or mistaken belief on the part of the speaker. Our claim is that the process of eliminating such cases as possible interpretations forces the hearer to make additional belief ascriptions which the speaker can rely on to communicate conversational implicatures.

The paper is structured as follows: in Sections 2 and 3 we describe ViewGen, a belief model which explicitly represents belief ascription and inference, and its use in plan recognition. In Section 4, we attempt to distinguish clear cases of mistaken belief and deception. Our claim is that for a conversational implicature to be recognised, such interpretations must be eliminated. Therefore, a belief set which eliminates both cases is presented as a condition for interpreting the utterance as an implicature. In Section 5, we show that the intermediate steps involved in elimination can be used to infer the intended content of the implicature.

Finally, in Section 6, we compare our theory with other approaches.

## 2 ViewGen

ViewGen [Wilks et al., 1991] is a dialogue understanding system which reasons about the attitudes of other agents in nested belief structures. It does this by *ascription* - i.e. assuming that attitudes held in one attitude box can be ascribed to others. There are two main methods of ascription - default ascription and stereotypical ascription. Default ascription applies to common attitudes which ViewGen assumes that any agent will hold and also ascribe to any other agent unless there is contrary evidence. Stereotypical ascription applies to stereotypical attitudes which ViewGen assumes apply to instances of a particular stereotype e.g.: ViewGen ascribes expert medical knowledge to doctors. Stereotypes can also apply to types of dialogue. For example, knowledge elicitation goals can be ascribed to agents involved in information seeking dialogues. Where as default ascriptions can be applied to any agent, stereotypical ascription requires the precondition of some trigger for the stereotype to be applied. For example, the use of a particular speech act stereotypically assumes the holding of certain attitudes.

ViewGen is capable of recognising and using a set of speech acts [Lee and Wilks, 1996b]. The acts are specified as plan operators which are used to plan what is said and recognise dialogue plans from the user. For example, informing is specified as:

**Inform(Speaker,Hearer,Proposition)**
**Preconditions: goal(Speaker,bel(Hearer,Proposition)) bel(Speaker,Proposition)**

where the predicates goal and bel refer to goals and beliefs.

Rather than specify the effects of an act, there are separate ascription rules for the speaker and hearer of any act:

Update on the Speaker's belief set
 For every condition C in a dialogue act performed:
  *ascribe(Speaker, Hearer, bel(C))*

Update on the Hearer's belief set
 For every condition C in a dialogue act performed:
  *ascribe(Hearer, Speaker, C)*

This account differs from standard accounts of computational speech acts (e.g. [Allen, 1983]) in that the underlying operation of ascription takes over implicitly many of the conditions and effects that are specified explicitly in other approaches. For example, the Inform act given above creates the following belief ascriptions on the part of the hearer:

*bel(Hearer, goal(Speaker,bel(Hearer,Proposition)))*
*bel(Hearer, bel(Speaker,Proposition))*

This highlights an important point about the definition of speech acts. Whether an Inform act succeeds depends on the communication of the speaker's mental state and not the actual intended effect of the act i.e. getting the hearer to believe P. This is the case in all speech acts - rather than define acts by effect, they are defined in terms of their conventional preconditions and their success or failure depends on the communication of these belief based preconditions. This allows acts to be planned by their minimal, context free effects. Further effects due to the context, e.g. whether the Inform act actually updates the hearer's personal belief of the communicated proposition, are separate from the act and modelled as further inferences or ascriptions in the dialogue plan. For example, there is an Accept_belief plan operator which allows the hearer to accept a speaker's belief if 1) there is no contrary evidence; 2) the speaker is a reliable source of information (i.e. an expert in the domain in question etc.). This allows a clear representation of how the conventional meaning of an utterance interacts with the context to produce a non-conventional effect such as in the case of an indirect speech act.

## 3 Plan recognition in ViewGen

ViewGen explicitly plans every belief ascription from the effects of an utterance to the desired interpretation, which results in an explicit representation of the speaker's intentions which can be used for further inference.

ViewGen plans its communicative goals using a partial order clausal link planner [McAllester and Rosenblatt, 1991]. The planner is also used to understand utterances by plan recognition. Given a communicative act, ViewGen attempts to generate a plan involving the act which results in goals which it can ascribe to the speaker. Ascribable goals are stereotypical goals for what is known about the user or context. Understanding can be seen as a form of ascription where ViewGen has to reason about which beliefs, goals and intentions are ascribed to the modelled agent. Further discussion of ViewGen's speech act representation and ascription are given in [Lee and Wilks, 1996a,Lee and Wilks, 1996b].

Plan recognition proceeds as follows: given an utterance, its minimal meaning is derived from its surface form. An utterance's minimal meaning is the set of communicated attitudes associated with the speech act of the utterance. The planner then attempts to generate a plan which connects the utterance with one or more ascribable goals. If the plan derived is inefficient i.e. fails to achieve the ascribed goals in the most direct manner possible, ViewGen assesses the speaker's plan by re-planning from the initial context to the ascribed goal. It then ascribes additional goals to the planning agent to explain the recognised plan's divergence from the "optimal plan". The optimal plan is considered to be the shortest the planner can generate from the context to the goal state. For example, the speaker might choose a particular dialogue plan to avoid mentioning a particular topic or diverge so that he or she can accomplish another goal implicitly.

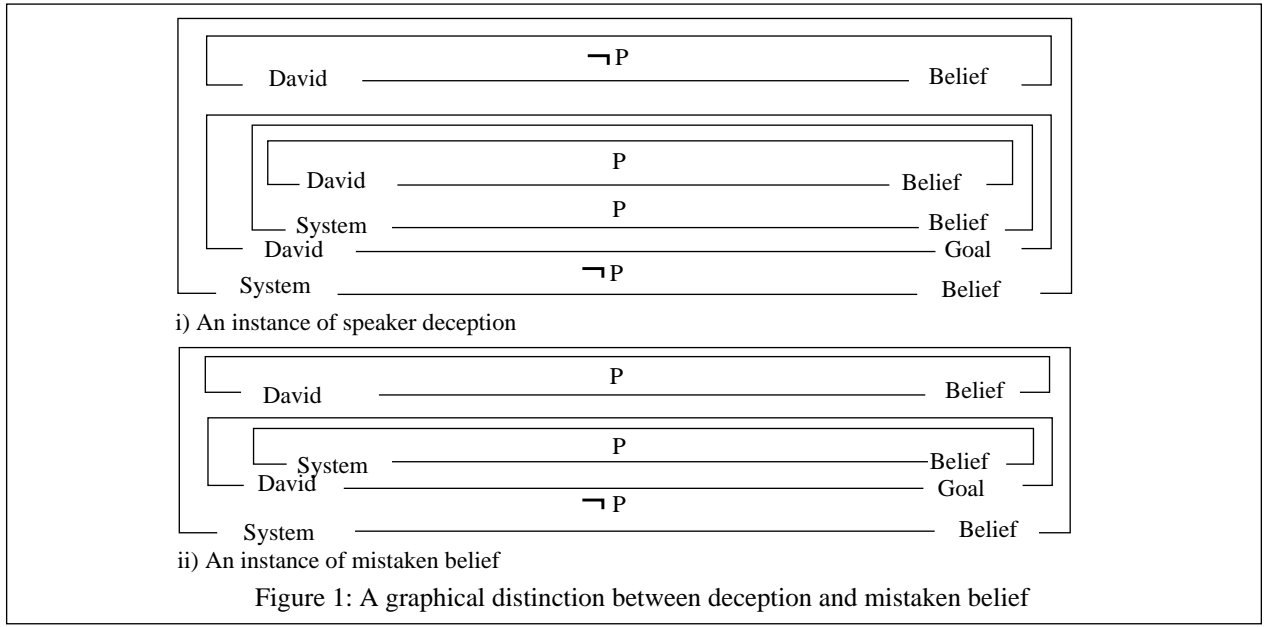

Figure 1: A graphical distinction between deception and mistaken belief

## 4 Cases of deception and mistaken belief

An act of deception occurs when a speech act is performed intentionally to communicate a false proposition. Deception can be identified when the inferred mental attitudes behind the utterance disagree with the hearer's prior beliefs about the speaker. However, cases of deception need to be distinguished from cases of mistaken belief. This is problematic since mistaken beliefs are also recognised when the speaker's reported mental attitudes are inconsistent with what the hearer believes to be true. For example, consider the following utterance:

**(2). David: Mark isn't here.**

If P represents not(present(Mark)) then David has performed the following Inform act:

**Inform(David, System, P)**

As noted above, the Inform act stereotypically provides justification for the following ascriptions by the System:

**bel(System, bel(David, P))**
**bel(System, goal(David, bel(System, P)))**

So far, in constructing the resultant belief environment, we have assumed that there are no prior assumptions about the speaker's belief space. However, prior beliefs strongly affect the interpretation of any utterance's meaning. The following two subsections demonstrate the use of such beliefs in recognising cases of deception and mistaken belief.

### 4.1 Deception by the speaker

Consider, the case where the following beliefs are present prior to the utterance:

**bel(System, not(P))**
**bel(System, bel(David, not(P)))**

That is, the System believes not P and believes that David also holds this belief. When the Inform act is performed then the latter belief blocks the ascription of the System believing that David believes P since it constitutes contrary evidence. However, the ascription of David having a goal of getting the System to believe P succeeds.

In cases of deception, it is clear that the speaker has a goal of getting the hearer to believe that the speaker holds a communicated proposition, so that in the case above, David is attempting to convince the System that he holds the proposition. That is:

**goal(David, bel(System, bel(David, P)))**

This attitude can be ascribed in the same manner as any felicity condition so that the hearer's belief can be updated thus:

**bel(System, goal(David, bel(System, bel(David, P))))**

This is a paradigm case of deception and can be generalised into the following rule:

Update on the Hearer's belief set in the case of deception

For every fraudulently communicated condition C in dialogue act performed:

*default_ascribe(Hearer, Speaker, goal(Speaker, bel(Hearer,C)))*

The application of this rule and the conventional update rule would result in the belief set shown in Figure 1(i).

### 4.2 Mistaken belief on the part of the speaker

Clearly, cases of deception and honest mistaken belief must be distinguished. Consider, by contrast; the case where the following beliefs are present:

**bel(System, not(P))**
**bel(System, bel(David, P))**

That is, the System believes not(P) but believes that David holds the contrary belief. When the Inform act is performed, there is no contradiction in the conventional ascriptions associated with informing. The updated belief set is shown in Figure 1(ii). However, because of the System holding the contrary belief not(P), the communicated belief cannot be accepted by the System as a primitive belief by the use of an Accept_belief operator.

### 4.3 Cases of deception and mistaken belief

In the above two subsections, we have distinguished paradigm cases of deception and mistaken belief based on the beliefs of the system of the speaker's actual beliefs. When an utterance is recognised, the conventional felicity conditions can be ascribed to the speaker. If these conditions are not believed to be true by the system then if there is evidence of deception, i.e., the System explicitly believes that the speaker holds a belief contrary to the utterance then the deception update rule can be applied and plan recognition based on this deception attempted.

However, it is more common for the system to hold no explicit belief on the speaker's personal beliefs with respect to the utterance prior to communication. Such a situation would occur if the System only holds its own personal belief about the status of P i.e.,

**bel(System, not(P))**

Without any further information about the speaker's beliefs, it is clearly not possible to guess correctly whether the speaker is being deceptive or truthful but mistaken. However, it is possible that the speaker is being neither deceptive or mistaken but attempting to communicate a conversational implicature. The understanding of such is given in section 5.

### 4.4 Plan recognition with mistaken beliefs and deception

ViewGen's nested representation of attitudes allows a consistent method with dealing with conflicting beliefs [Ballim and Wilks, 1991]. It is simple to represent conflicting belief environments held by agents and to construct and add nested beliefs using ascription. Since both types of ascription are blocked by contrary evidence, it is possible to reason about which mental attitudes an agent holds even if the agent shows evidence of a mistaken belief or is being deceptive.

In the case of a mistaken belief, ViewGen is able to plan from the utterance and the context of the speaker's belief environment as usual. Additional beliefs can be ascribed into the environment as required as long as they are not blocked by the mistaken belief.

Like any utterance, an act of deception is motivated to achieve goals by the speaker. It is, therefore, possible to use plan recognition to reason about the speaker's dialogue plan. Such cases require that the System assumes that the speaker believes the deception will be successful. For example, in the above case, the belief set ascribed to John in Figure 1(i) can be used to generate a plan with John's belief set as the initial context via the ascribed goal of deception to some ascribable goal.

However, in general, the above definition of deception is too strict. In most situations, the system will not have the nested beliefs required to recognise genuine cases of deception. The recognition of deception and its motivation is clearly a form of keyhole recognition i.e. unaided and unintended by the speaker and difficult at best. However, as shown in Section 5, such a definition can be used to eliminate the possibility of deception and mistaken belief in cases where the speaker intends the hearer to recognise the utterance as a conversational implicature. Ostensible but intentionally recognisable infelicities are commonly used to communicate information by cooperative plan recognition.

### 5. Making inferences from infelicities

In this section, we briefly describe an extension of the theory of speech act understanding described in section 3. Our theory is based on the idea that recognising and distinguishing that the utterance is not a case of deception or mistaken belief and, therefore, a conversational implicature results in a set of ascribed beliefs which constitute the meaning of the implicature.

Given such an utterance, it is first essential to recognise that the speaker is attempting an act of cooperative communication rather than being deceptive or holding a mistaken belief. Such recognition involves eliminating the possibility of the speaker intending deception or holding a mistaken belief. In the case of eliminating a case of mistaken belief, the System must have an explicit evidence that the speaker believes the contrary belief. That is, if P is the mistaken belief then:

**bel(System, bel(Speaker, not(P)))**

If the speaker intends that the System interpret his or her utterance pragmatically then they must believe that this contrary evidence is believed by the System prior to performing the speech act i.e.

**bel(Speaker, bel(System, bel(Speaker, not(P))))**

In the case of any genuine act of deception by the speaker, he or she must believe that the intended deception was successful i.e. not detected by the hearer as such. As argued above, detection of deception requires, if P is the communicated proposition, that the System holds the following attitude:

**bel(System, bel(Speaker, not(P)))**

Therefore, if the speaker is to be sure that the utterance cannot be recognised as a deception, he or she must hold the following belief:

**bel(Speaker, bel(System, bel(Speaker, not(P))))**

This is the same belief as required for preventing the interpretation of mistaken belief. For the System to recognise that the utterance is neither a case of deception or mistaken belief, it must therefore nest this belief to eliminate either possibility i.e.:

**bel(System, bel(Speaker, bel(System, bel(Speaker,**

**not(P)))))**

If the System can construct such a belief by ascription after the speaker's utterance then it can recognise that the speaker is attempting to communicate an implicature by appearing to flout a felicity condition.

As noted in section 2, ViewGen constructs belief environments as required, representing explicit beliefs only when necessary. In such a case, the act of distinguishing the utterance as pragmatic, typically results in a series of belief ascriptions as the cases of deception and mistaken belief are eliminated and the utterance recognised as cooperative. Because agents are able to simulate each other as nested environments, it is possible for a speaker to plan the hearer making such belief ascriptions in order to understand his or her utterance and intend such ascriptions as the non-conventional, context sensitive aspect of the meaning of the utterance.

For example, reconsider the dialogue exchange in (1). Suppose initially, the System holds the following attitudes:

**goal(System, bel(Simon, deleted(System, sub_directory)))**
**bel(System, deleted(System, sub_directory))**
**bel(System, deleted(_, sub_directory) -> not(happy(David)))**
**bel(System, bel(Simon, deleted(_, sub_directory) -> not(happy(David))))**

After the System's act of informing that the subdirectory has been deleted, the following beliefs can be stereotypically ascribed by the System to Simon:

**bel(System, bel(Simon, bel(System, deleted(System, sub_directory))))**
**bel(System, bel(Simon, goal(System, bel(Simon, deleted(System, sub_directory)))))**

After recognising Simon's act of informing that David will not be happy, the System can attempt to ascribe the following attitudes based on the felicity conditions of the speech act:

**bel(System, bel(Simon, happy(David)))**
**bel(System, goal(Simon, bel(System, happy(David))))**

However, the first belief is blocked due to the contrary evidence in Simon's belief environment since it contains the belief that deleting a sub-directory will result in David not being happy and the belief that the System has deleted the sub-directory. These propositions licence the inference that Simon believes David is, in fact, not happy, i.e.

**i. bel(System, bel(Simon, not(happy(David))))**

Since this blocks the ascription of the first felicity condition of Simon's speech act, the system must then attempt to reason whether the utterance is a case of mistaken belief, deception or a use of implicature. For the latter, the System must attempt to derive:

**bel(System, bel(Speaker, bel(System, bel(Speaker, not(P)))))**

The System can simulate Simon ascribing to the System the belief that deleting the sub-directory will make David unhappy to derive:

**ii. bel(System, bel(Simon, bel(System, deleted(_, sub_directory) -> not(happy(David)))))**

and then the System can simulate Simon simulating the System using this proposition to infer that David will not be happy, i.e.

**iii. bel(System, bel(Simon, bel(Sytem, not(happy(David)))))**

Inferring beliefs **i** and **iii**, allows the System to eliminate the possibility of an act of deception or mistaken belief by Simon. This matches the requirements for recognising that the speaker must be intending an implicature by his utterance. It is therefore necessary to attempt to understand what the content of the implicature is.

This interpretation depends not on the conventional effects of the speech act, since its felicity conditions are believed to have been disobeyed; but rather the context sensitive effects caused by the hearer attempting to recognise that the speaker is attempting to be pragmatically communicative. In attempting to interpret the pragmatic content of Simon's utterance, the System therefore must derive a plan which involves at least one of these effects as a sub-goal of a goal which can be stereotypically ascribed.

Such a plan is shown in Figure 2. From the belief **iii**, the System derives that Simon believed that the System now believes that David is not happy by the System's action of deleting the subdirectory. This is a stereotypical goal which can be ascribed to Simon and therefore, an acceptable interpretation.. The goal ascription blocks the second felicity condition of the original inform act of holding a goal to get the System to believe the opposite. Therefore, Simon's utterance can be interpreted as a use of irony where the intended meaning is opposite to the conventional meaning.

## 6 Related work

Perrault [Perrault, 1990] presents a theory of speech acts based on default logic. Using a definition of belief and intention expressed in the logic, he defines a set of default rules for speech act performance and considers the application of such rules to cases of deception and mistaken belief. Because of their default nature, prior inferences can be retracted to deal with such problematic cases. Perrault's work can be seen as a basis for our own. The idea of default rules for the speech act effects is similar in many respects to our concept of ascription. However, our work extends to link the elimination of deception and mistaken belief to the inference of conversational implicature.

Pollack [Pollack, 1990] presents an account of plan recognition which deals with the possibility of mistaken belief on the part of the speaker by ascribing a set of beliefs associated with his or her plan plus a set of meta level beliefs explaining the role the first set of beliefs play in the speaker's plan. Pollack's idea of ascribing a set of beliefs based on the intentions contained in the recognised plan is similar to our own though she does not discuss how conversational implicatures can be dealt with or consider deception.

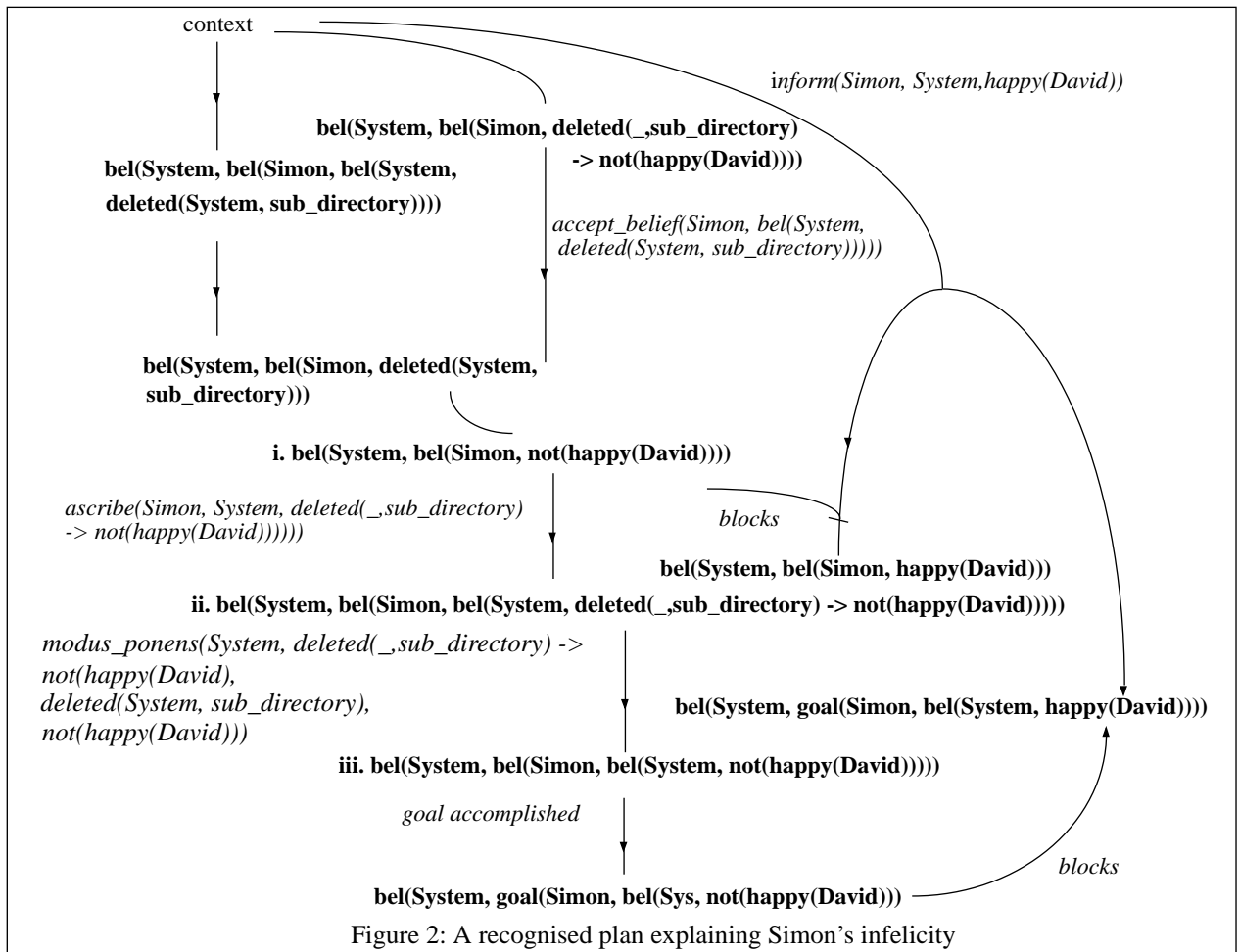

Figure 2: A recognised plan explaining Simon's infelicity

Utsumi [Utsumi, 1996] presents a unified theory of irony which includes a concept of insincerity. He argues that there are three properties: allusion, pragmatic insincerity and emotive attitude. Given two of these properties, the third can be inferred and the utterance judged ironic. His concept of pragmatic insincerity refers to the utterance either flouting a speech act's felicity conditions or a pragmatic principle such as the maxim of quantity [Grice, 1975], the politeness principle [Leech, 1983], or the principle of relevance [Sperber and Wilson, 1986].

Utsumi's concept of pragmatic insincerity is clearly related to ours. However, he presents no way of determining whether the pragmatic insincerity is intended or due to deception or mistaken belief. Utsumi's work is also more general than ours in that we only consider felicity conditions whereas Utsumi appeals to a wide range of pragmatic theories for his concept of sincerity. However, principles such as politeness and relevance are too underspecified to be computationally usable. It is an open research question whether such principles are required.

## 7 Discussion and Conclusions

In the sections above, we have shown that the belief attitudes often found in non-cooperative dialogue can be used in cooperative dialogue to recognise and understand a class of conversational implicature.

However, it is clear that our definition of deception is too strict for use in all but the most simple cases of non-cooperative deceptive dialogues. Clearly, sophisticated uses of deception involve more than the flouting of the felicity conditions of a single speech act. For example, a speaker may attempt to establish a proposition earlier in the dialogue to validate the conditions of the main deception or attempt to mislead the hearer by seizing the initiative of the dialogue to draw it away from the actual subject of the deception.

The main use of these definitions is in distinguishing such cases from where the speaker is attempting to communicate an implicature in cooperative dialogue. In this paper, we have avoided actually trying to define the class of implicature and instead defined conditions for what is not in the class. Whether this is theoretically adequate is an open question.